%% file: emnlp-ijcnlp-2019.tex
\documentclass[11pt,a4paper]{article}
\usepackage[hyperref]{emnlp-ijcnlp-2019}
\usepackage{times}
\usepackage{url}
\usepackage[T1]{fontenc}
\usepackage{mathtools}
\usepackage{amsfonts}
\usepackage{bbm}
\usepackage{times}
\usepackage{latexsym}

\usepackage{url}

\aclfinalcopy 

\title{Fine-grained Sentiment Analysis with Faithful Attention}
\author{ 
  Ruiqi Zhong, 
  Steven Shao, 
  Kathleen McKeown
  \\
  Department of Computer Science, Columbia University \\
  { \{rz2383, ys2833\}@columbia.edu}
  , kathy@cs.columbia.edu
  }

\date{}

\begin{document}
\maketitle

\input{Introduction.tex}

\input{Related.tex}

\input{Data.tex}

\input{Method.tex}

\input{Metrics.tex}

\input{cmp_w_bl.tex}

\input{Results.tex}

\input{Discussion.tex}
\bibliography{emnlp-ijcnlp-2019}
\bibliographystyle{acl_natbib}
\clearpage
\appendix

\input{Appendix.tex}

\end{document}

%% file: Introduction.tex
\def\hide#1{\iffalse #1 \fi}
\def\todo [#1]#2{{\color{red} \textbf{For #1}: #2}}
\begin{abstract}
    While the general task of textual sentiment classification has been widely studied, much less research looks specifically at sentiment between a specified source and target. To tackle this problem, we experimented with a state-of-the-art relation extraction model. Surprisingly, we found that despite reasonable performance, the model's attention was often systematically misaligned with the words that contribute to sentiment. Thus, we directly trained the model's attention with human rationales and improved our model performance by a robust 4$\sim$8 points on all tasks we defined on our data sets. We also present a rigorous analysis 
    of
    the model's attention, both trained and untrained, using novel and intuitive metrics. Our results show that untrained attention does not provide faithful explanations; however, trained attention with concisely annotated human rationales not only increases performance, but also brings faithful explanations. 
    Encouragingly, a small amount of annotated human rationales suffice to correct the attention in our task.
\end{abstract}

\section{Introduction}

While the problems of detecting targeted sentiment and sentiment aspect have been well-studied, there has been less work on the detection of sentiment relations in text. 
In genres such as Twitter or other social media sites where targeted sentiment is applied, the source of the sentiment is almost always the author. In more formal genres such as news, however, the author frequently specifies who feels a certain sentiment about a particular target, usually using words that provide cues about the sentiment (which we term {\em rationales}) as the sentences in Table~\ref{tab:examples} show. 
In early work on the analysis of sentiment relations, ~\citet{Ruppenhofer2008} note that in meetings and blogs several sources may be present, discussing different issues. 
A system that only identifies positive or negative sentiment towards a target misses the more limited scope of the sentiment intended by the author.  
The task of finding the sentiment of opinion relations given the source and target closely resembles the task of relation extraction as the system must predict the value of the sentiment between two entities. 

\def\bt#1{\boxed{\text{#1}}}
\begin{table}
  \centering
  \begin{tabular}{l}\hline
  \textbf{Sentences} \\ \hline ``\bt{He} \underline{endorsed} a \bt{proposal} for targeted \\ sanctions \underline{against} \bt{Mugabe}.'' \\ ``\bt{The sale} \underline{infuriated} \bt{Beijing}, which \underline{regards} \\ \bt{Taiwan} an integral part of its territory.'' \\ \hline \hline
  \textbf{Relations} \\ \hline \small{He $\xrightarrow[\text{endorsed}]{+}$ proposal, proposal $\xrightarrow[\text{against}]{-}$ Mugabe} \\ \small{The sale $\xrightarrow[\text{infuriated}]{-}$ Beijing, Beijing $\xrightarrow[\text{regards}]{+}$ Taiwan }
  \end{tabular}
  \caption{Typical sentences in our dataset, which includes entity (boxed) and relation (as arrows) annotations with human rationales (underlined).}
  \label{tab:examples}
\end{table}

In this paper, we present a method for extracting opinion relations that features a technique for training attention to focus on human rationales, along with a series of quantitative metrics to evaluate the quality of the model's attention. 
Our approach builds on a state-of-the-art relation extraction method~\cite{zhang2017position} and augments it with a loss function based on KL-divergence between the model's attention and the human rationales in our training data. 
We evaluated our system on two separate annotated corpora which carry different kinds of sentiment: the Multi-Perspective Question Answering (MPQA) corpus~\cite{wilson2008fine} which has been extensively used in sentiment analysis, and the Good-For Bad-For (GFBF) corpus~\cite{deng2013benefactive} which contains sentences expressing whether a person or event is good or bad for another person or event. 
We also developed our system for two separate scenarios, one in which it must distinguish between pairs of entities that have sentiment relations and pairs that do not, and another in which it receives only entity pairs that do participate in sentiment relations where the goal is to determine polarity.

We compared our approach against a wide range of relation extraction systems showing a significant absolute gain in performance of 8 points. 
We experimented with different quantities of human-annotated rationales and show that only a small number of human rationales was enough to yield half of the gain in performance. 

Finally, we used two novel and intuitive metrics, ``probes-needed" and ``mass-needed" and a simple LIME \cite{ribeiro2016should} based method to verify that our model does attend to words that it relies on to make the prediction, and hence brings \textit{faithful attention} - whether it truly reflects how the model makes prediction.
In contrast, our experiments reveal that on one of our tasks, the standard use of attention in \cite{zhang2017position} does not focus on the words that the model relies on to make the prediction. 

Our contributions include:
\begin{itemize}
    \item A novel model for opinion relation extraction that significantly outperforms other recent relation extraction approaches by incorporating human rationales.
    
    \item Rigorous and intuitive quantitative metrics to evaluate model's attention.
    
    \item An encouraging finding that only a small number of concisely annotated human rationales is needed to improve performance and provide faithful explanation.

\end{itemize}

%% file: Related.tex
\section{Related Work}
This work touches a wide range of domains, including sentiment analysis, relation extraction and attention mechanisms.  

\paragraph{Sentiment Analysis}
Previous neural methods for targeted sentiment analysis have developed modified versions of attention that take extended context surrounding the target words into account ~\cite{LiuandZhang2017,Changetal2018}. Neither of these take human rationales into account nor do they consider the source of sentiment. More recent work on targeted sentiment analysis (without source), like our work, points to problems in the accuracy of using attention and proposes instead the use of a transformative network model~\cite{Lietal2018}. 

\citet{Yang&Cardie2013} use sequence labeling to jointly identify sentiment relations including source as well as target and in later work they develop  an LSTM for the same task~\cite{katiyar2016investigating}, but they do not identify sentiment polarity and thus, while the focus overlaps with ours, it is not directly comparable.
Other related research uses probabilistic soft logic (PSL)~\cite{deng-wiebe:2015:EMNLP}, along with multiple separate models for inferring source, target, sentiment span and polarities and merges the results using different PSL approaches. 
Their pipelined approach contains heuristics, however, and is difficult to replicate. 

\paragraph{Relation Extraction}
There has been a wide range of neural network architectures designed for the task of relation extraction, e.g., Convolutional Neural Networks with attention mechanism \cite{wang2016relation}  augmented by dependency structures \cite{huang2017improving, zhang2018graph}, Tree Neural Networks \cite{socher2013recursive} with LSTM units \cite{tai2015improved, zhu2015long, miwa2016end}, and  LSTMs over dependency trees \cite{xu2015classifying}.
Our model is an extension of \citet{zhang2017position}, and we compared our model against \citet{xu2015classifying}, \citet{tai2015improved} and \citet{wang2016relation}. 
We show that our approach that incorporates human rationales outperforms all of these by a wide margin.

\paragraph{Attention Mechanisms}
Usually attention is learned in an unsupervised fashion, but several recent approaches use supervised techniques to improve attention with human rationales.
For instance, \citet{mi2016supervised} used word alignment to guide machine translation and \citet{liu2017exploiting} used argument keywords to improve event detection.
A more recent sophisticated approach~\cite{bao2018deriving} trains a model that maps from human rationales to machine attention which are then used for target tasks.
However, none of these works examine a model's attention quantitatively and systematically.

Our attention evaluation is closest to a very recent paper by \citet{jain2019attention}, which also provided a rigorous analysis on the faithfulness of model's attention.  
They found that attention weights are poorly correlated with the influence of each token approximated by the leave-one-out method; additionally, they point out that incorporating human rationale is a promising future direction in its related work section.
In comparison, through evaluating ``mass/probes needed", the two novel metrics specifically designed for this sentiment analysis task, we also found that untrained attention does not bring faithful attention.
Excitingly, we are able to move one step forward and our experiment shows that incorporating human rationales can correct this problem under some cases - and a small amount of them suffice. 

%% file: Data.tex
\section{Data} \label{Data}
\paragraph{General Formulation}
In our formulation, each sentence annotation consists of (i) a source text span (an entity), (ii) a target text span (an entity/event), (iii) a rationale span, and (iv) a label. 
For example, in the sentence, \emph{``I respect my collaborator''}, ``I'' is the source, ``my collaborator'' is the target, ``respect'' is the rationale describing the relation from the source to the target, and the sentiment label is positive. 
More precisely, let sentence $S$ be a word sequence $[w_{0}, w_{1}, w_{2} ,\ldots ]$; source text span $s = (s_{start}, s_{end})$ s.t. $[w_{s_{start}}, w_{s_{start+1}}, \ldots, w_{s_{end - 1}}]$ is the text mention of the source; similarly target text span $t = (t_{start}, t_{end})$; and rationale span $c = (c_{start}, c_{end})$. 
In the above example, $w_{0}$ = ``I'', $s = (0, 1), c = (1, 2), t = (2,4)$.
Our task is to predict the sentiment label given $S$, $s$ and $t$, where $s$ and $t$ are ground-truth entity spans as in the standard formulation of relation extraction problems. 

In one setting for our model, the ground-truth entity pair is known to have a relation and the model must classify its polarity, e.g., as positive or negative. Not every pair of entities/events has a relation, however. In the above example, the converse sentiment, from ``my collaborator'' towards ``I'', is unknown.  Thus, we consider a second setting, detailed in Section \ref{recnetbaseline}, that requires our model to also \emph{identify} relations between ground-truth entities. 
For both data sets, we performed 5-fold cross validation with 90$\%$ of the data. Within each fold, 65$\%$ of the documents were assigned to training, 15$\%$ development, and 20$\%$ test. The remaining 10\% of the data is held out for comparing our new method's performance against the baseline's.

\paragraph{MPQA 2.0} The Multi-Perspective Question Answering 2.0 dataset \cite{wilson2008fine} annotates entities as well as relations between them, with those classified as ``direct-subjective'' and ``attitude'' corresponding to sentiment. 
These relation annotations include offsets that indicate rationale spans, and their ``polarity'' attributes, which could be one of \textit{positive}, \textit{negative} or \textit{neutral}, can be used to determine labels.
Details on the annotation scheme can be found in \citet{wilson2008fine}. We pre-processed the data into the format shown in Table \ref{tab:examples}, 
and created the data points as described in the previous paragraph. This resulted in $\sim$800 positive, $\sim$900 neutral, $\sim$750 negative, and $\sim$17000 $\emptyset$-labeled data points for training on each fold.

\paragraph{GFBF}
The GoodFor/BadFor corpus \cite{deng2013benefactive} is similar to MPQA2.0, but instead annotates whenever an entity positively or negatively affects another. This dataset is still relevant for sentiment because the relations are semantically similar, and there are also only two categories, positive and negative (as opposed to other relation extraction data sets, such as TACRED \cite{zhang2017position} and SemEval-2010 Task 8 \cite{hendrickx2009semeval}). A separate benefit of this dataset is its shorter rationale spans compared to MPQA, making it simpler to evaluate attention (discussed later).  
Processing for this corpus is similar to MPQA, resulting in $\sim$300 ``goodFor", $\sim$400 ``badFor" and $\sim$2300 $\emptyset$-labeled training points on each fold.

\paragraph{Example Annotations} We include some examples in the Appendix Section \ref{sampledata}.

%% file: Method.tex
\section{Methods} 
In this section, we describe (i) our model's architecture and the baselines we compare against and (ii) our method of training attention with human rationales. We include our code for implementation in Supplementary Materials.\footnote{The software and processed dataset will be released under the MIT License upon acceptance.}
\subsection{Baseline Experiments} \label{recnetbaseline}
Our experiments center around the model developed by \citet{zhang2017position}, which comprises a Long Short Term Memory network (LSTM) and attention mechanism (abbreviated as AttnLSTM). 
We also implemented a Convolutional Neural Network \cite{kim2014convolutional} with multi-level attention \cite{wang2016relation} and an LSTM over the Shortest Dependency Path \cite{xu2015classifying}, and a Tree-LSTM \cite{tai2015improved}. 
These systems have all been developed to identify relations between input entities and thus, they are tested on data that includes gold standard entities. 
We follow this approach in our work. 

\def\h{\mathbf{h}}
\def\x{\mathbf{x}}
\def\p{\mathbf{p}}
\def\attn{\hat{A}}
\def\softmax{\operatorname{softmax}}
\def\loss{\mathcal{L}}
\def\crossentropy{\operatorname{\textsc{CrossEntropy}}}
\def\cel{\mathcal{L}_{clf}} 
\def\E{\mathbb{E}}
\def\W{\mathbf{W}}
\def\e{\mathbf{e}}
\def\forrz#1{{\color{red}\textbf{For Ruiqi:} #1}}
\def\fc{\textit{fc}}
\def\L{\mathcal{L}}
\def\Lattn{\L_{attn}}

Here we give a quick overview of the AttnLSTM model.
The original AttnLSTM represents each entity-annotated sentence with a plain Bi-LSTM's output $\h_i$ over a sequence of word vectors $\x_i$:
\begin{equation}
    \{\h_1, \ldots, \h_{|S|}\} = \text{Bi-LSTM}(\{\x_1, \ldots, \x_{|S|}\})
\end{equation}
The model uses randomly initialized position embeddings, $\p_i^s$ and $\p_i^t$, to identify source and target respectively. For example, $\p_i^s$ is a vector in an embedding matrix $\mathbf{P}$ indexed by the displacement $p_i^s$ between $i$ and the source span $s$:
\begin{equation}
    p_i^s = \min(i-s_{start},0) + \max(0,i-s_{end})
\end{equation}
$\p_i^t$ is computed analogously, using the same embedding matrix $\mathbf{P}$. $\p_i^s$ and $\p_i^t$ are then used, along with $\h_{i}$ and the final hidden state $\h_{q}$, to determine the attention weights $\attn$. That is,
\begin{equation}
    \e_{i} = \tanh(\W_h\h_i + \W_q \h_q + \W_s\p_i^s + \W_t\p_i^t)\\
\end{equation}
\begin{equation}
    u_i = \mathbf{v}^\top \e_{i}
\end{equation}
\begin{equation}
    \attn = \softmax(\{u_1, \ldots, u_{|S|}\})
\end{equation}
where $\W_h, \W_q, \W_s, \W_t, \mathbf{v}$ are trained parameters.

The final sentence representation used for classification is then a weighted sum, $\mathbf{z} = \sum_{i=1}^n \attn(i)\h_i$, to be fed into a fully connected layer with softmax activation for classification; 
i.e. $y = softmax(W_{z}\mathbf{z})$ is the predicted probability distribution over labels, where $W_{z}$ is a fully connected layer with trained parameters.

As in the original paper, we utilize multi-channel augmentation and word dropout; 
in our experiments with this model, the $\x_i$ is a concatenation of pretrained word embedding, part-of-speech embeddings and sentiment class embeddings. 
We also mask the source and target span to UNK and randomly set $6\%$ of the words to UNK to simulate dropout.
Two additional trainable vector parameters are used to replace, or \emph{mask}, the embeddings for words in the source and target span, respectively. 

The classification loss $\loss_{clf}$ is the cross-entropy between prediction $y$ and label $l$.
We consider more loss components in the next section. 

In the setting where the model must identify relations, each pair of annotated entities/events within a sentence becomes a data point, with some pairs annotated as having ``no-relation'', abbreviated $\emptyset$. 
The resultant dataset is heavily skewed (since $N$ entities within a sentence will lead to roughly $N^2$ pairs, with all but a few labelled $\emptyset$), so we used under-sampling to balance the label distribution.

\subsection{Leveraging Human Rationales} \label{leveragecues}
A human annotator may focus on certain cue words while determining sentiment; therefore, these human rationales can be used to guide the model's attention. 
We use these human rationales to extend the AttnLSTM, which many standard relation extraction models, as well as our base AttnLSTM, would be unable to utilize during training.

Our method to incorporate human rationales is to directly train the model's attention $\hat{A}$. 
Since $\hat{A}$ is a probability distribution over $|S|$ positions, we define a human's ``ground truth attention" $A$ also as a probability distribution that is uniform over the human rationales, or uniform over every word if the label is $\emptyset$. That is,
\begin{equation} \label{rationaledef}
    A(i) = 
    \begin{cases}
        \frac{1}{|S|} & \text{if } l = \emptyset \\
        \frac{1}{c_{end} - c_{start}} & \text{else if } c_{start} \leq i < c_{end}\\
        0 & \text{otherwise}
    \end{cases}
\end{equation}
The model's attention is made to match ``human attention" through a new KL-divergence loss
component $\Lattn = KL(A||\hat{A})$, thus making the overall objective $\L = \L_{clf} + \lambda_{attn}\Lattn$, where $\lambda_{attn}$ is a tunable hyperparameter.

\textbf{Predicting Rationales} An alternative approach is a standard multi-task learning framework, where the model must additionally assign the probabilities that each word would be annotated as within the span of the rationale and hence add a corresponding loss component $\L_{r}$, a tunable hyper-parameter $\lambda_{r}$ to balance the main task of predicting the label with the auxiliary task of identifying human rationales.
Since it is a standard and popular approach, we include the details in the appendix section \ref{pred-cues}. 
We tune the multi-task learning parameter $\lambda_{r}$ thoroughly for each task, while keeping $\lambda_{attn}$ fixed after initial experiments). 
This is to negate any tuning-related bias in performance \cite{lipton2018troubling} in favor of our novel formulation when comparing it to the multi-task learning baseline on cross-validation test sets.

\textbf{``Limited Rationales"} As human rationale annotations are expensive, it is natural to ask "When human rationales are unavailable, how much would it help if the researchers manually annotated some?".
To answer this question, we simulated scenarios where only a limited number of human rationales are annotated by training with only a small subset of human rationales in the dataset.
To make notations easier we index data points by superscript $n$. We sample a subset of relations in our dataset on which to apply our attention loss, and rescale $\Lattn^n$ appropriately.
In particular, if $M$ is a random $\gamma$-fraction of points in our dataset, we define a subsampled loss
\begin{equation}
    \Lattn'^{n} = \begin{cases}
        0 & \text{ if } l^{n} \neq \emptyset \text{ \& } n \notin M\\
        \Lattn^n / \gamma & \text{ if } l^n \neq \emptyset\text{ \& }n \in M \\
        \Lattn^n & \text{ if } l^n = \emptyset \\
    \end{cases}
\end{equation}
\begin{equation}\L^n = \L^n_{clf} + \lambda_{attn}\Lattn'^{n}\end{equation}
so that sampled relations have losses scaled up, non-sampled ones zeroed, and $\emptyset$ unchanged.


%% file: Metrics.tex
\section{Metrics}
In this section we describe how we evaluate model performance and attention. 
We provide both high level explanations and formal mathematical definitions for the model's attention metrics. 

\subsection{Prediction Accuracy Metrics} 
In the setting where the model merely needs to classify relations as positive/neutral/negative or goodFor/badFor, the label classes are fairly balanced, so we measure the model's accuracy. In contrast, when the model also has to identify relations (i.e., data points may be labelled $\emptyset$) we measure F-score because of the class imbalance.
\def\argmax{\operatorname{argmax}}

\subsection{Faithfulness Metrics} \label{attnmetrics}
Here we give our novel metrics to evaluate the model's attention: probes-needed and mass-needed. 
On a high level, these two metrics represent: if a person wants to find the token that influences the model's prediction the most, but is only given the attention weights on each position for his/her reference, how much time/``cost" does it take to find the token?

We first determine the most influential token using the approximation LIME (Locally Interpretable Model-agnostic Explanations) \cite{ribeiro2016should}. 
LIME scores each word by comparing the model's confidence score on its original prediction between the unmodified sentence and a sentence with the current word masked as UNK, so that the ``most important word'' to the model is the one that would lower the confidence the most if masked. 
Formally, let $y$ be the confidence score assigned to its predicted label; $S_{-i}$ be the sentence $S$ with the $i^{th}$ being replaced by UNK; $y_{-i}$ be the confidence score with the input sentence $S_{-i}$. 
Then our most influential token is at the position $i = \argmax_j y - y_{-j}$.

\textbf{Probes-needed} refers to the number of  words we need to probe (look at) based on the model's attention to find the most important word. Specifically, this is the ranking of the rationale word if the words were sorted descending by attention weight assigned by the model. In this case, the model that largely ignores the rationale word would have a much higher ``probes-needed'' since the word would be near last in this weight ranking.

Formally, let the sentence length be $|S|$ and $i$, the most important position, be defined as $i = \argmax_j A(j)$ (taking the first $i$ if tied), where $A$ is the human rationale defined in equation \ref{rationaledef}. 
Now we consider a model attention $\hat{A}$ and find the index $i$ according to $\hat{A}$.  
Therefore,
\begin{equation} \label{probesneeded}
    \text{probes-needed}(\hat{A}) = 1 + \sum_{j | \hat{A}(j) > \hat{A}(i)}1
\end{equation}

\textbf{Mass-needed} is even more fine-grained. 
If the words are sorted as in probes-needed, mass-needed is the total probability mass (i.e. attention weight) on the words ranked above the rationale word. This distinguishes cases when, for example, two models both assign the rationale word second place, but one a much more distant second (so its ``mass needed'' would be higher since more mass is on an incorrect first). Formally,

\begin{equation} \label{massneeded}
    \text{mass-needed}(\hat{A}) = 1 + \sum_{j | \hat{A}(j) > \hat{A}(i)}\hat{A}(j)
\end{equation}

\citet{jain2019attention} evaluated the faithfulness of attention mechanism by calculating the Kendall $\tau$ correlation between the attention weights and individual token influences $y - y_{-j}$, and they acknowledged that irrelevant features may add noise to the correlation measures.
In contrast, we only focus on the most influential words. 
Such a decision is well-suited for this sentiment analysis task, where only a few tokens are dominantly important in determining the label.

\subsection{Plausibility Metrics} \label{plausible}
Attention weights seem to explain the models' behavior because they sometimes match the key words that human uses to make a decision. 
However, an explanation \textit{plausible} to human eyes does not imply that the interpretation is faithful. 
In this paper, we also evaluate the plausibility of the attention weights.
Since on GFBF the average rationale length is less than 2 words, we can also use the probes/mass-needed metrics to evaluate how much time/``cost" the attention weights take to find a rationale token.
In contrast, rationales on MPQA are long (> 4 words), so we evaluate plausibility through human crowd sourcing experiments.
We used a crowd-sourced platform, Figure-8, to let humans ultimately judge how well each attention distribution ``explains'' a particular sentiment.~\footnote{This was approved by an IRB with exempt status under protocol number AAAS0867, each worker consented electronically by choosing to participate after being given the instructions.} 
We sampled a set of 40 data points from each of the 5 folds in which the rationale-trained and base models both predicted the correct sentiment relation with confidence over $0.5$ to filter out noisy predictions, and showed workers attention visualizations of the two models, asking them to compare the quality of the two (paying 61 cents for 8 questions). 
In particular, we asked (i) whether either of the attention visualizations is a sensible explanation and (ii) if both are sensible, which one is preferred, and by how much. We ultimately judged an attention mechanism to explain a relation better than another on a data point if either (i) this one was sensible but not the other, or (ii) both were sensible, but this one was preferred. Figure \ref{Figure-8} in the Appendix shows an example Figure-8 question.

%% file: cmp_w_bl.tex
\begin{table*}[t]
\centering
\begin{tabular}{ccccc}
\hline
Model & MPQA Include-$\emptyset$ & MPQA Exclude-$\emptyset$ & GFBF Include-$\emptyset$ & GFBF Exclude-$\emptyset$ \\
\hline
AttnCNN & 20.3 & 38.7 & 34.5 & 53.6\\
\hline
TreeLSTM & 29.9 & 56.1 & 46.4 & 70.6 \\
\hline
SDP & 34.2 & 60.7 & 60.1 & 81.0\\
\hline
AttnLSTM & 32.7 & 62.4 & 60.4 & 80.4\\
\hline
Pred-rationales & 34.9 & 63.0 & 61.4 & 85.1\\
\hline
Trained-attn & \textbf{37.6}(+4.9) & \textbf{68.7}(+6.3) & \textbf{65.4}(+5.0) & \textbf{88.7}(+8.3)\\
\hline
\end{tabular}
\caption {\label{baselines} F-score (for Include-$\emptyset$ columns) and accuracy (for Exclude-$\emptyset$ columns) for various baselines and our method. AttnCNN is the CNN model by \citet{wang2016relation}; TreeLSTM is a tree model with LSTM units by \citet{tai2015improved}; SDP is the Shortest Dependency Path model by \citet{xu2015classifying}; AttnLSTM is our baseline model by \citet{zhang2017position}. Pred-rationales is a popular multi-task learning method for leveraging rationales, and trained-attn is our novel method for training the attention. Both of them are applied to AttnLSTM, and are described in Section \ref{leveragecues}. Improvements of Trained-attn over non-trained attention, shown in parentheses, are statistically significant with $p < 0.002$. Likewise, improvements over Pred-rationales have $p < 0.02$.}
\end{table*}

%% file: Results.tex
\section{Results}
\label{sect:results}
We first show that our novel method for training attention brings significant improvements and outperforms a wide range of competitive baselines. 
Next, we establish that a small number of (potentially expensive) human rationale annotations suffices to bring half of our empirical gain. 
Finally, we quantitatively reveal that in some tasks the attention of the original AttnLSTM exhibits unexpected behaviors.
On GFBF our method provides a simple fix and trained attention brings both faithful explanations and plausibility;
however, on MPQA our method only improves plausibility, but the attention is not faithful. 

\subsection{Comparison with Baselines}
Table \ref{baselines} shows the performance of each model.
Trained attention brings a significant improvement on F-score over  untrained attention (4$\sim$ 8 points); additionally, it improves all the label-wise precision, recall and F-scores on both MPQA and GFBF (\ref{breakdown}). 
It also outperforms the popular multi-task learning method (Pred-rationales) in all tasks, even though $\lambda_{rationale}$ has been tuned much more thoroughly than $\lambda_{attn}$. 
\footnote{As a sanity check against overtuning, we compared the performance of AttnLSTM vs. Trained-attn on the held-out test set. We found a 3.5-point improvement on MPQA Include-$\emptyset$, 7.9 on MPQA Exclude-$\emptyset$, 2.8 on GFBF Include-$\emptyset$, and 9.2 on GFBF Exclude-$\emptyset$, improvements similar to that on the cross-validation test sets.}

\subsection{Varying the Number of Rationales}
\begin{figure}[h]
    \centering
    \includegraphics[width=\linewidth]{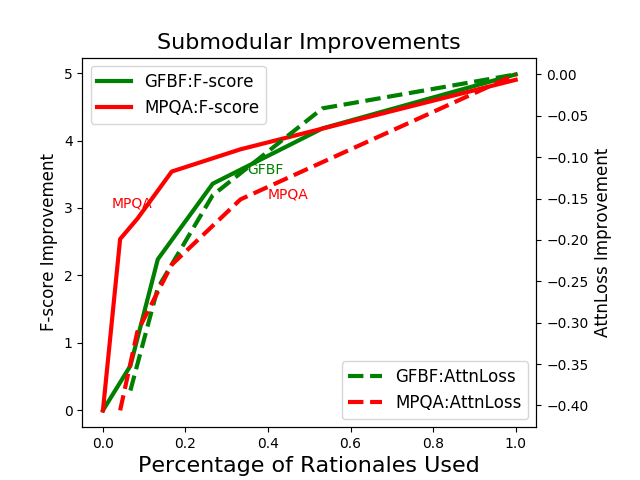}
    \caption{On the left/right hand side is F-score Performance/attention loss improvement (y-axis) vs. using different percentages of human-annotated rationales (x-axis). We calculate the difference from attention loss when training with all rationales. To draw the plot, for MPQA, we sample 100, 200, 400, and 800 rationales (corresponding to 4\%, 8\%, 16\%, and 33\% of all rationales); and for GFBF, 50, 100, 200, and 400 rationales (7\%, 13\%, 27\%, and 53\%).}
    \label{fig:varyrationale}
\end{figure}

These results describe the performance of a model trained as briefly introduced in Section \ref{leveragecues}. 
Figure \ref{fig:varyrationale} plots this performance against the fraction of data points with human rationales annotation.
As expected, the more rationales, the better the performance.
Notably, the improvement is sub-modular: the marginal improvement per rationale decreases as more rationales are used, and so the first few rationales end up being the most helpful. 
Just 100 rationales (4\% of all rationales in MPQA, or 7\% in GFBF) brings about 2.5 points of improvement. Attention loss, defined in section \ref{attnmetrics}, follows a similar increasing and submodular trend, for both datasets. 

Even more surprisingly, we found that while a model could develop a reasonable attention mechanism by itself with just a good amount of data without rationales, training on a few rationales can `nudge' the attention in the right direction, resulting in significant gains in attention performance. 
In a separate experiment, we trained a model on just 200 data points and all their rationales, and this performed worse than training on all the data points but a smaller fraction of rationales -- 50 rationales (4\%) on GFBF, and 100 (7\%) on MPQA. Therefore, fewer rationales can be more effective if there is enough other data.

\subsection{Problems Corrected on GFBF} \label{faithfulness-results}

\definecolor{medium}{RGB}{102, 160, 255}
\definecolor{light}{RGB}{226, 237, 255}
\definecolor{dark}{RGB}{0, 46, 122}
\def\m#1{\color{white}\colorbox{medium}{#1} \color{black}}
\def\d#1{\color{white}{\colorbox{dark}{#1}} \color{black}}
\def\l#1{\colorbox{light}{#1}}

\begin{figure}[h]
\begin{tabular}{r|l}
Untrained & \bt{Socialized medicine} will hurt \\ & \bt{\d{Medicare}} and is \l{too} expensive. \\ \hline 
Trained & \bt{Socialized medicine} will \d{hurt} \\ & \bt{Medicare} and is too expensive.
\end{tabular}
\caption{\label{fig:badattn}A typical data point where the untrained model attends to the target, and the trained attends to the key word for the relation (in this case, between ``Socialized medicine'' and ``Medicare'').}
\end{figure}

For GFBF models, trained attention weights improve significantly for both explanation faithfulness and plausibility, evaluated by probes/mass-needed. 
If the attention is untrained, on average we need to probe 8 positions (probes-needed) to find the  ``most important word'', the word that the model relies on the most (faithful explanation), or the rationale word (plausibility). 
However, adding just 100 rationales for training causes a drastic reduction in probes-needed to around 2, almost as high as training with all rationales. 
We observe the same trend for mass-needed as well ( Table~\ref{GFBFExplain}). 
An important observation is that a random baseline is expected to have a mass needed of 0.5, which the non-trained attention system significantly exceeds. 
This suggests that it develops an attention mechanism that actually systematically fails to attend to the correct position. 

\begin{table}[h]
\begin{center}
\begin{tabular}{cccc}
\hline 
Plausibility & Probes-needed & Mass-needed \\ 
Non-trained  & 8.00 & 0.80 \\ 
100 rationales & 1.73 & 0.128\\ 
Fully trained  & \textbf{1.55} & \textbf{0.111}\\
\hline 
Faithfulness & Probes-needed & Mass-needed \\ 
Non-trained & 8.46/8.71 & 0.78/0.73 \\ 
100 rationales & 2.06/3.24 & 0.12/0.20\\ 
Fully trained & \textbf{2.02/2.68} & \textbf{0.10/0.16}\\ \hline
\end{tabular}
\caption{\label{GFBFExplain} Attention metrics on models trained with varying numbers of rationales. The lower the Probes/Mass -needed, the better the attention faithfulness.  We report the cases where the prediction is correct / wrong separately.}
\end{center}
\end{table}
\subsection{Plausibility$\neq$ Faithful Explanation on MPQA} \label{pvsf}
We report the plausibility (Section \ref{plausible}) and faithfulness (Section \ref{attnmetrics}) of the attention weights produced by the model trained on MPQA.

We compare the plausibility of trained vs. non-trained attention in our crowd sourcing experiment, and humans in general find the trained attention weights to be more plausible. 
Among the $\sim$700 collected evaluations about the relative quality of the two attention system outputs, non-trained attention was deemed sensible $87.9\%$ of the time, and trained attention $93.3\%$. 
Overall, non-trained attention was judged to be better $18.0\%$ of the time and trained-attention $34.7\%$ of the time ($47.3\%$ being  draws), a result that is significant with $p < 6\cdot 10^{-8}$. 
Hence, trained attention is more plausible on MPQA.

Nevertheless, the attention weights are not ``faithful" explanations. 
It requires 16.98 probes and 0.688 (> 0.5) mass to find the most influential word, which exceeds the random baseline.
We conclude that for trained attention model on MPQA, attention explanation is mostly plausible but poorly faithful.


%% file: Discussion.tex
\section{Discussion}
\label{sect:discussion}

\paragraph{Plausibility vs. Faithful Explanations}
An important distinction throughout our paper is the three ``interpretations" of a prediction: (1) the words we (humans) rely on when judging a sentiment relation, (2) the words a model relies on in making a prediction, and (3) the attention weights of each word.
Only the attention weights are immediately visible, since rationales only approximate the distribution of words humans rely on to make the predictions, and the model may hide a pathway separate from attention for making sentiment predictions.
As shown in section \ref{pvsf}, explanation faithfulness and plausibility might not be both present.

Typically if the model is good enough, it should naturally rely on what the human relies on; however, this is not always the case, i.e. as discussed by \citet{rajpurkar2018know} and \citet{mudrakarta2018did}. 
In such a setting, all three quantities could differ, but our results provide evidence that forcing the model's attention to align with human rationales (plausibility) has the side-effect of uniting attention with the model's hidden word attribution (faithfulness) on the GFBF corpus.

\paragraph{Repairing Attention} 
As discussed by \citet{bao2018deriving}, attention mechanisms degrade when there is not enough training data. 
Our results corroborate this claim (refer back to Table~\ref{GFBFExplain}), showing that it might be even worse: rather than being noisy, the model's attention is \textbf{systematically wrong} on some folds when trained on small datasets like GFBF. 
As shown in Table \ref{GFBFExplain}, untrained attention sometimes brings so little faithfulness that it attends to the rationale word less frequently than a random baseline, suggesting that the model is systematically assigning weights to words that it does not rely on to make the prediction.

On the GFBF corpus, our method provides a simple fix. 
As shown in Table \ref{GFBFExplain}, trained attention brings faithful explanations, even with just a few rationales. 
However, 
we see that it does not work under all circumstances:
on MPQA the trained attention weights do not provide faithful explanations. 
We suspect the reason is that the rationales annotated on MPQA are long and contain ``redundant words".
Consider the following example:
``The visit is widely expected to be centered on [China]$_{\text{source}}$'s [dissatisfaction over]$_{\text{rationale}}$ [the Taiwan issue]$_{\text{target}}$.''
The model might not need to use the word ``over" in its prediction, thus introducing noise into our induced ground truth attention.
This implies that more ``concisely annotated" rationales might be helpful: only the minimum set of words that determines the sentiment is needed.
For example, in figure \ref{fig:badattn}, the word ``hurt" alone describes the relation between the source and target, and other words such as ``will" should be excluded.

Since our technique does not use task-specific engineering and only requires human-annotated rationales and the model's attention, it can be easily applied to other types of relation extraction tasks with concise human-annotated rationales to improve both the performance and attention faithfulness.
Nevertheless, maintaining a faithful attention is \textit{inherently} hard for other tasks: the interpretation that attention mechanism induces is inevitably a weighted sum of relevant words, obliterating the complex syntactic, grammatical, and compositional relation between them, which are important for most of the NLP tasks beyond sentiment analysis. 
Therefore, our methodology is more likely to be applied when the most difficult part of the task involves finding and disambiguating the few tokens that matter to the label, rather than disambiguating complex syntactic relations.

%% file: Appendix.tex
\def\notice#1{{\color{blue}\textbf{Note:} #1}}

\def\source#1{[#1]$_{s}$}
\def\target#1{[#1]$_{t}$}
\def\rationale#1{[#1]$_{r}$}
\def\sentence{\textbf{Sentence: }}
\def\badfor{\textbf{Label}: {\color{red} \textbf{Badfor}}}
\def\goodfor{\textbf{Label}: {\color{green} \textbf{Goodfor}}}
\def\negative{\textbf{Label}: {\color{red} \textbf{Negative}}}
\def\positive{\textbf{Label}: {\color{green} \textbf{Positive}}}
\def\neutral{\textbf{Label}: {\color{gray} \textbf{Neutral}}}

\section{Supplementary Materials}
\subsection{Sample Data} \label{sampledata}
Here we include some example annotations from the pre-processed corpus. $[]_{s}$ is the source span, $[]_{t}$ the target span and $[]_{r}$ the rationale span. 

\subsubsection{MPQA2.0}
\sentence After Putin 's \target{statement} \source{they} \rationale{rubbed their palms at length}.
\\\positive
\\
\\\sentence The situation in which \source{Putin} \rationale{has agreed} \target{to open former Soviet military airfields to American armed forces in three (former Soviet) Central Asian countries} could be regarded as dramatic .
\\\positive
\\
\\\sentence \source{Putin} \rationale{has no greater desire than} \target{to present to the West the Chechen independence movement as a chapter of `` international terror}.'' He did not succeed with that so far .
\\\positive
\\
\\\sentence \source{I} \rationale{didn't particularly want} to \target{come here} and when I came here everybody noticed that I was really, really sad because I loved being at Enderly Park.
\\\negative
\\
\\\sentence Talking to IRNA, \source{Ivashev} \rationale{expressed regrets over} \target{the fact that the people in some parts of the world are indifferent toward the expansionist policies of the United States and do not proceed to condemn them} .
\\\negative
\\
\\\sentence Talking to IRNA, \source{Ivashev} expressed regrets over the fact that the people in some parts of the world are indifferent toward \rationale{the expansionist policies of \target{the United States} and do not proceed to condemn them}.
\\\negative
\\
\\\sentence Talking to IRNA, Ivashev expressed regrets over the fact that \source{the people in some parts of the world} are indifferent toward the expansionist policies of the United States and \rationale{do not proceed to condemn} \target{them}.
\\\neutral
\\
\\\sentence While depending heavily on his rural support, \source{Mugabe} told thousands of urbanizers he \rationale{would focus on} \target{housing and job creation} if re-elected .
\\\neutral
\\
\\\sentence Mbeki was, however, already on the record as saying that \source{the Commonwealth} also had \target{other issues} to \rationale{consider}.
\\\neutral
\subsubsection{GFBF}
\sentence Now more than ever is the time to remind him that he 's had his day and \target{the policies} \source{he} \rationale{implemented} - from the stimulus to Obamacare to new financial regulation - not only haven't helped; they've hurt.
\\\goodfor 
\\
\\\sentence And early indications about ObamaCare's implementation via new regulations suggest \source{this law} will \rationale{validate} \target{its critics' dire predictions}
\\\goodfor
\\
\\\sentence Apparently, \source{bureaucrats} will do what the ObamaCare law didn't to \rationale{ensure} \target{that government takes over health care}.
\\\goodfor
\\
\\\sentence By \rationale{limiting} \target{its repeal}, \source{Congress} unconstitutionally entrenched IPAB , preventing members of Congress from effectively representing their constituents
\\\badfor
\\\source{Those reductions} will further \rationale{diminish} \target{the number of Medicare providers} and/or reduce the quality of carein essence, creating precisely the de facto rationing of health-care services the bill supposedly prohibits
\\\badfor
\\
\\\sentence In addition to ignoring some existing problems in the health care system, \source{ObamaCare} will actually \rationale{worsen} \target{other problems for doctors}
\\\badfor

\subsection{Hyperparameters} \label{hyperp}
Here we describe our implementation of our model based on \cite{zhang2017position}. We used the 300-dimensional word vectors pretrained on GoogleNews \cite{pennington2014glove}. Each word had an additional 10 channels for a part-of-speech embedding (POS Tagging by nltk \cite{bird2009natural}), and another 10 for an embedding according to the word's sentiment score induced from a Reddit news community corpus, using SocialSent \cite{hamilton2016inducing}. The other parameters in \citet{zhang2017position}, including the hidden state dimension $|\h_i| = d$, position embedding dimension $d_p$, and latent dimension for computing attention $d_a$, were $140$, $35$, and $50$, respectively.

We tuned $\lambda_{r}$ on the test set of the first fold for each task, pick the best $\lambda_{r}$, and use it for all the other folds. In particular, we varied $\lambda_{r}$ to be [0.01, 0.025, 0.05, 0.1, 0.15, 0.2, 0.25, 0.3] and found 0.05 to be the best for MPQA including $\emptyset$, 0.25 for MPQA excluding $\emptyset$, 0.25 for GFBF including $\emptyset$, and 0.3 for GFBF excluding $\emptyset$.
We tuned $\lambda_{attn}$ as little as possible, since we wanted to make sure that it was our novel training method that brought empirical improvement, not the hyperparameter tuning. We experimented with $\lambda_{attn} = 0.3$ in our first attempt and it worked well enough, so we fixed it throughout the entire paper.

 \subsection{Figure 8 Example Question} \label{explainmetrics}

\begin{figure}[h!]\label{Figure-8}
    \centering
    \includegraphics[width=\linewidth]{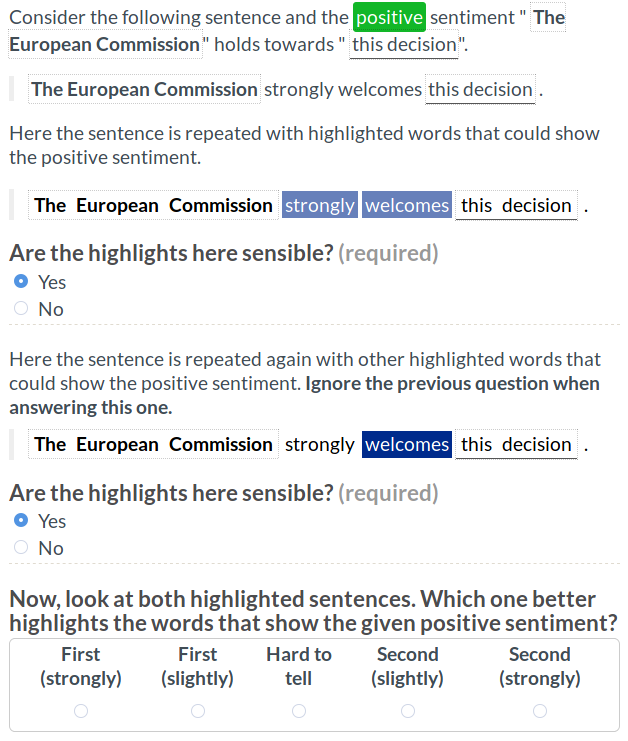}
    \caption{An example question shown to workers.}
    \label{Figure-8}
\end{figure}
 \label{interpretmetrics}
 
 \subsection{Predicting Cues Formulation} \label{pred-cues}
 Formally, the ground truth of rationale is defined as
\begin{equation}
    c_{i} = \hide{\mathbbm{1}[l \neq \emptyset \text{ and } c_{start} \leq i < c_{end}]}        \begin{cases}
        1 & l^{n} \neq \emptyset \text{ and } c_{start} \leq i < c_{end}\\
        0 & \text{otherwise}
        \end{cases}
\end{equation} 
and the model is augmented with a component $\hat{c}_i = \sigma(\fc_{r}^{T}(\e_i))$ that predicts $c_i$ by feeding the word representation $\e_i$ through a fully connected layer $\fc_{r}$ and then a sigmoid. $\L_{r}$; a cross-entropy loss between $\hat{c}_i$ and $c_i$, is used to train, thus making the overall objective $\mathcal{L} = \L_{clf} + \lambda_{r}\mathcal{L}_{r}$, where $\lambda_{r}$ is a tunable hyperparameter.
This allows the model to use the same amount of information as training attention, but does not train the model's attention directly.
 
 \subsection{Performance Breakdown} 
 Because of the large class imbalance towards $l^n=\emptyset$, we compute precision and recall with a kind of micro-average over data points that actually have, or are classified to have a relation.
\begin{equation}
    P(\text{Precision}) = \frac{|\{n \in N\,|\, y^{n} = l^{n} \neq \emptyset\}|}{|\{n \in N \,|\,y^{n} \neq \emptyset\}|}
\end{equation}
\begin{equation}
    R(\text{Recall}) = \frac{|\{n \in N\,|\, y^{n} = l^{n} \neq \emptyset\}|}{|\{n \in N \,|\,l^{n} \neq \emptyset\}|}
\end{equation}
\begin{equation}
    \text{F-score} = \frac{2PR}{P + R}
\end{equation} 
Since non-$\emptyset$ classes are nearly balanced, our metrics are practically identical to (< 1.5\% relative difference from) the macro-F score metrics.
Here we show the performance breakdown of trained attention by label and metrics (precision, recall and f-score), respectively for MPQA and GFBF. 
Number in the parentheses is the improvement over the original AttnLSTM model. 
 We observe that our approach with trained model's attention outperforms the original AttnLSTM for all metrics and labels. 
 \label{breakdown}
 \begin{table}[h!]
\begin{center}
\begin{tabular}{cccc}
\hline 
label & Precision & Recall & F-score \\ \hline
Negative & 36.4(+2.1) & 38.7(+2.5) & 36.9(+1.5) \\ \hline
Neutral & 30.8(+3.0) & 36.5(+4.6) & 32.9(+3.6)\\ \hline
Positive & 41.2(+9.0) & 44.7(+8.1) & 42.3(+8.2)\\ \hline
Average & 36.1(+4.7) & 40.0(+5.1) & 37.4(+4.5)\\ \hline
\end{tabular}
\caption{\label{MPQA-breakdown}Precision, recall and F-scores for each label on MPQA (including $\emptyset$), averaged across 5 folds. Each entry shows the performance with trained attention, with the improvement over AttnLSTM baseline in parentheses. Results are averaged over 5 folds, so $F \neq \frac{2PR}{P + R}$ in this table.}
\end{center}
\end{table}

\begin{table}[h!]
\begin{center}
\begin{tabular}{cccc}
\hline 
label & Precision & Recall & F-score \\ \hline
goodfor & 60.2(+5.9) & 66.9(+6.3) & 63.4(+6.7) \\ \hline
badfor & 63.4(+0.3) & 71.2(+7.4) & 66.8(+3.5)\\ \hline
Average & 61.8(+3.1) & 69.1(+6.9) & 65.1(+5.1)\\ \hline
\end{tabular}
\caption{\label{GFBF-breakdown}Performance breakdown on GFBF including $\emptyset$, similar to Table \ref{MPQA-breakdown}.}
\end{center}
\end{table}